\documentclass[a4paper,fleqn]{cas-dc}

\usepackage{multirow}
\usepackage[authoryear,longnamesfirst]{natbib}
\usepackage{array}
\usepackage{float}
\usepackage{siunitx}
\usepackage{stfloats} 
\usepackage{nidanfloat}
\usepackage{gensymb}
\newcolumntype{P}[1]{>{\centering\arraybackslash}p{#1}}

\def\tsc#1{\csdef{#1}{\textsc{\lowercase{#1}}\xspace}}
\tsc{WGM}
\tsc{QE}
\tsc{EP}
\tsc{PMS}
\tsc{BEC}
\tsc{DE}

\begin{document}
\let\WriteBookmarks\relax
\def\floatpagepagefraction{1}
\def\textpagefraction{.001}

\shorttitle{Through-Foliage Tracking with Airborne Optical Sectioning}

\shortauthors{Nathan, Kurmi, Schedl, and Bimber}

\title [mode = title]{Through-Foliage Tracking with Airborne Optical Sectioning}                      
\author[1]{Rakesh John Amala Arokia Nathan}[orcid=0000-0001-5534-0182]
\ead{rakesh.amala_arokia_nathan@jku.at}
\credit{Software, Data curation, Writing - Original draft preparation}

\author[1]{Indrajit Kurmi}[orcid=0000-0001-7065-0509]
\ead{indrajit.kurmi@jku.at}
\credit{Software, Data curation, Writing - Original draft preparation}

\author[2]{David C. Schedl}[orcid=0000-0002-7621-3526]
\ead{david.schedl@fh-hagenberg.at}
\credit{Software, Data curation}

\author[1]{Oliver Bimber}[orcid=0000-0001-9009-7827]
\ead{oliver.bimber@jku.at}
\credit{Conceptualization of this study, Methodology, Writing - Original draft preparation}
\cormark[1]
\cortext[cor1]{Corresponding author}

\affiliation[1]{organization={Johannes Kepler University Linz},
    addressline={Altenberger Straße 69}, 
    city={Linz},
    postcode={4040}, 
    country={Austria}}
    
\affiliation[2]{organization={University of Applied Sciences Upper Austria},
    addressline={Softwarepark 11}, 
    city={Hagenberg Austria},
    postcode={4232}, 
    country={Austria}}

\begin{abstract}
Detecting and tracking moving targets through foliage is difficult, and for many cases even impossible in regular aerial images and videos. We present an initial light-weight and drone-operated 1D camera array that supports parallel synthetic aperture aerial imaging. Our main finding is that color anomaly detection benefits significantly from image integration when compared to conventional raw images or video frames (on average 97\% vs. 42\% in precision in our field experiments). We demonstrate, that these two contributions can lead to the detection and tracking of moving people through densely occluding forest. 
\end{abstract}

\maketitle
\section{Introduction}
With Airborne Optical Sectioning (AOS,  \cite{kurmi2018airborne,kurmi2019statistical,kurmi2019thermal,kurmi2020fast,kurmi2021combined,kurmi2021pose,bimber2019synthetic,schedl2020airborne,schedl2020search,schedl2021autonomous}) we have introduced a wide synthetic aperture imaging technique that employs conventional drones to sample images within large areas from above forest. These images are computationally combined (registered to the ground and averaged) to integral images which suppress strong occlusion and make hidden targets visible. AOS relies on a statistical chance that a point on the forest ground is unoccluded by vegetation from multiple perspectives, as explained by the statistical probability model in \cite{kurmi2019statistical}. The integral images can be further analyzed to support, for instance, automated person classification with advanced deep neural-networks. In \cite{schedl2020search}, we have shown that integrating raw images before classification rather than combining classification results of raw images is significantly more effective when classifying partially occluded persons in aerial thermal images (92\% vs. 25\% average precision). In \cite{schedl2021autonomous}, we demonstrate a first fully autonomous drone for search and rescue based on AOS. The main advantages of AOS over alternatives, such as LiDAR or Synthetic Aperture Radar, are its real-time performance; its applicability to other wavelengths, such as far infrared for wildlife observations and search and rescue, or near infrared for agriculture and forestry applications; and its high spatial resolution. 

 \begin{figure}[!ht]
	\centering
		\includegraphics[width=\linewidth]{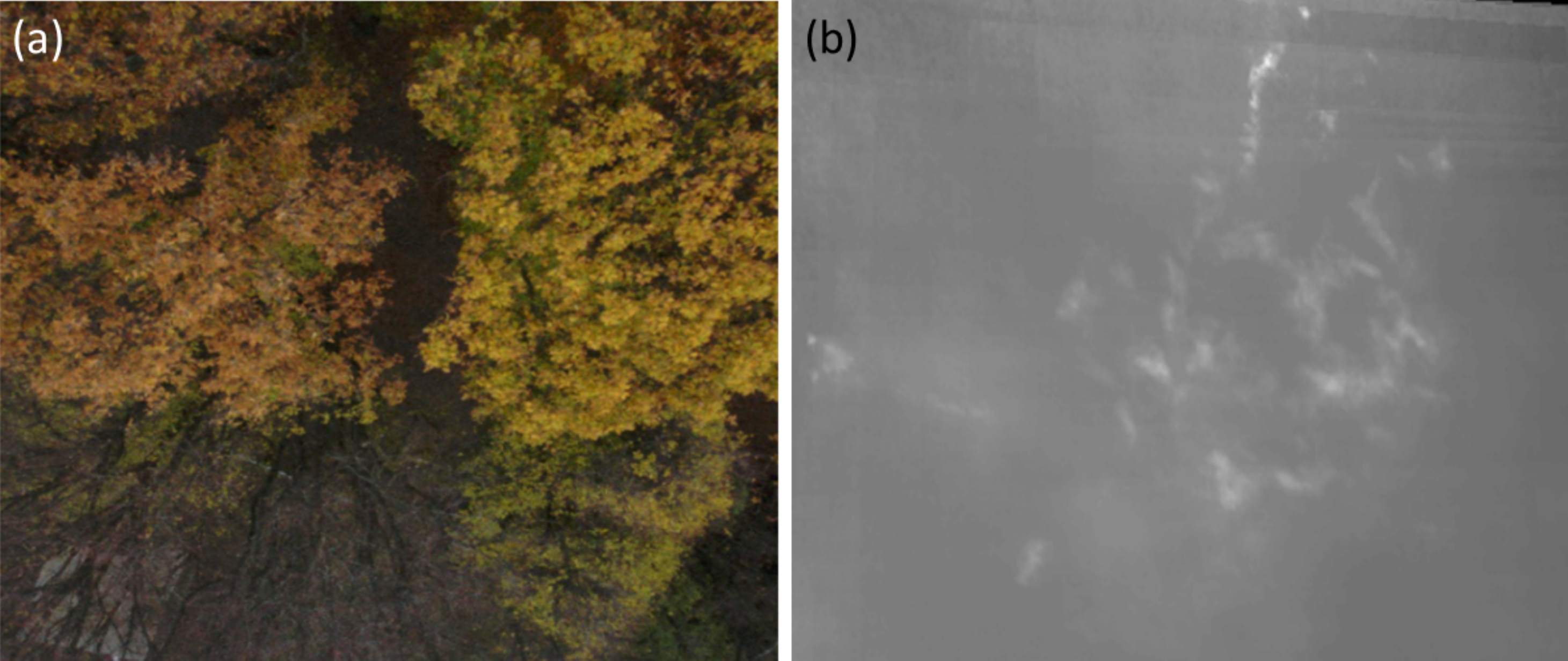}
	\caption{Sequential AOS sampling: A herd of moving red deer inside a densely occluding forest patch (a) caused strong motion blur in integral images (b). In this experiment, we sampled a synthetic aperture of 150mx150m with 500 thermal images in 12 minutes.}
	\label{FIG:0}
\end{figure}

Thus far, the sequential sampling nature of AOS, limits its applications to static targets only. Moving targets, such as walking people or running animals lead to motion blur in integral images that are nearly impossible to detect and track (cf. Fig. \ref{FIG:0}, for example). Applying AOS to far infrared (thermal imaging), as in \cite{schedl2020search,schedl2021autonomous}, restricts it to cold environment temperatures, while using it in the visible range (RGB imaging) as in \cite{kurmi2018airborne}, often suffers from too little sunlight penetrating through dense vegetation.  

In this article we make two main contributions: First, we present an initial light-weight (<1kg) and drone-operated 1D camera array that supports parallel AOS sampling. While 1D and 2D camera arrays have already been used for implementing various visual effects (e.g., \cite{Vaish04,vaish2006reconstructing,zhang2018synthetic,YangTao2016,Joshi2007,pei2013synthetic,YangTao2014}), they have not been applied for aerial imaging (in particular with drones) because of their size and weight. Second, we show that color anomaly detection (e.g., \cite{reed1990adaptive,ehret2019image}) benefits significantly from AOS integral images when compared to conventional raw images (on average 97\% vs. 42\% in precision). Color anomaly detection is often used for automatized aerial image analysis in search and rescue applications (e.g., \cite{Morse2012,Agcayazi2016,Weldon2020}) because of its independence to environment temperature (in contrast to thermal imaging). However, it fails in presence of occlusion. We demonstrate, that these two contributions can lead to the detection and tracking of moving people through dense forest.     
\section{Integral Imaging with Airborne Camera Arrays}

\begin{figure}[!t]
	\centering
		\includegraphics[width=\linewidth]{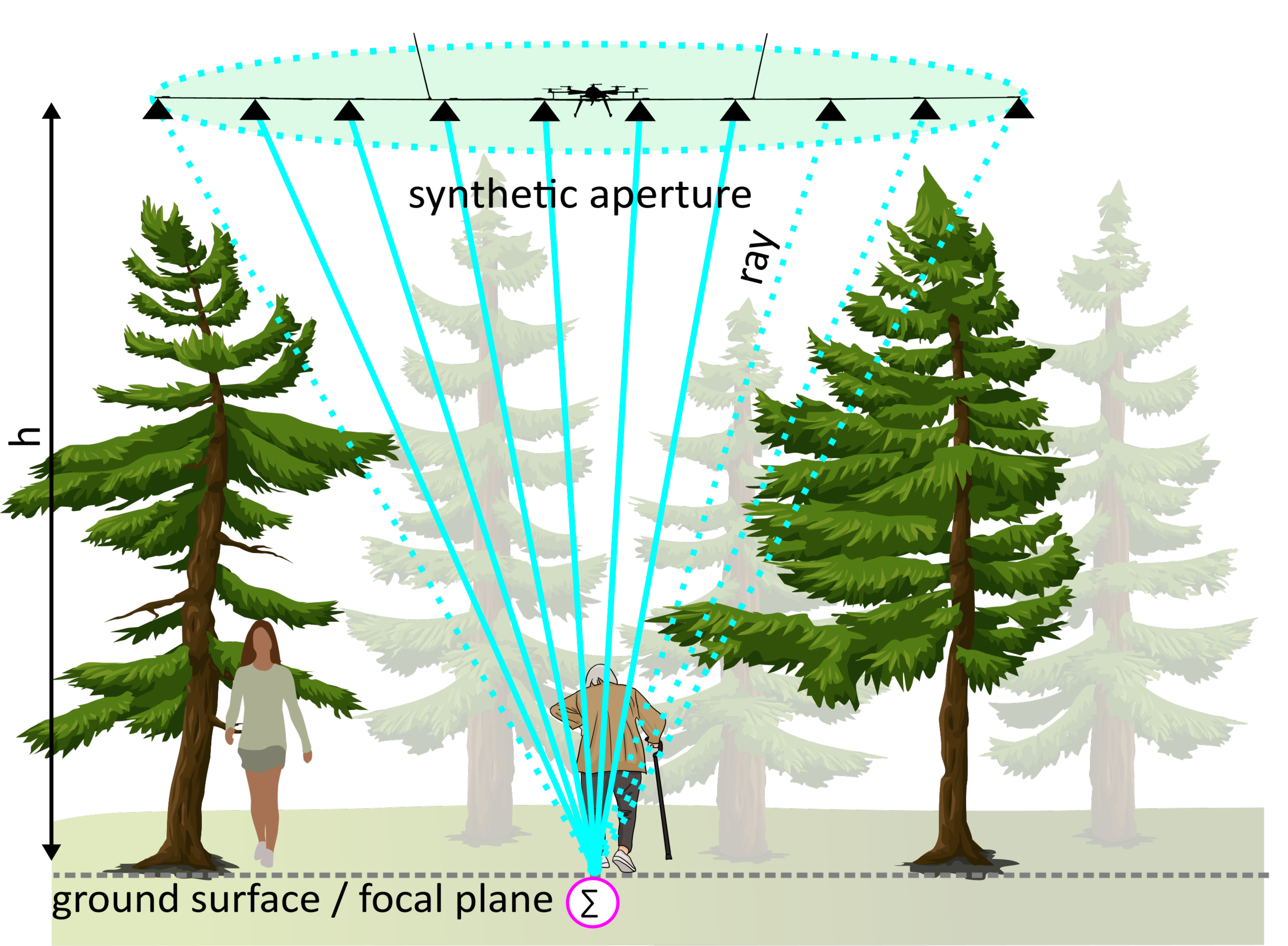}
	\caption{Airborne Optical Sectioning (AOS) captures multiple images within a synthetic aperture that, in this work, is defined by the size of a camera array. For occlusion removal, both --occluded and unoccluded rays (dashed and solid lines respectively)-- are integrated for a common target point on the ground surface or on an approximating focal plane (resulting in one pixel of the integral image).}
	\label{FIG:1}
\end{figure}
 \begin{figure*}[h]
	\centering
		\includegraphics[width=\linewidth]{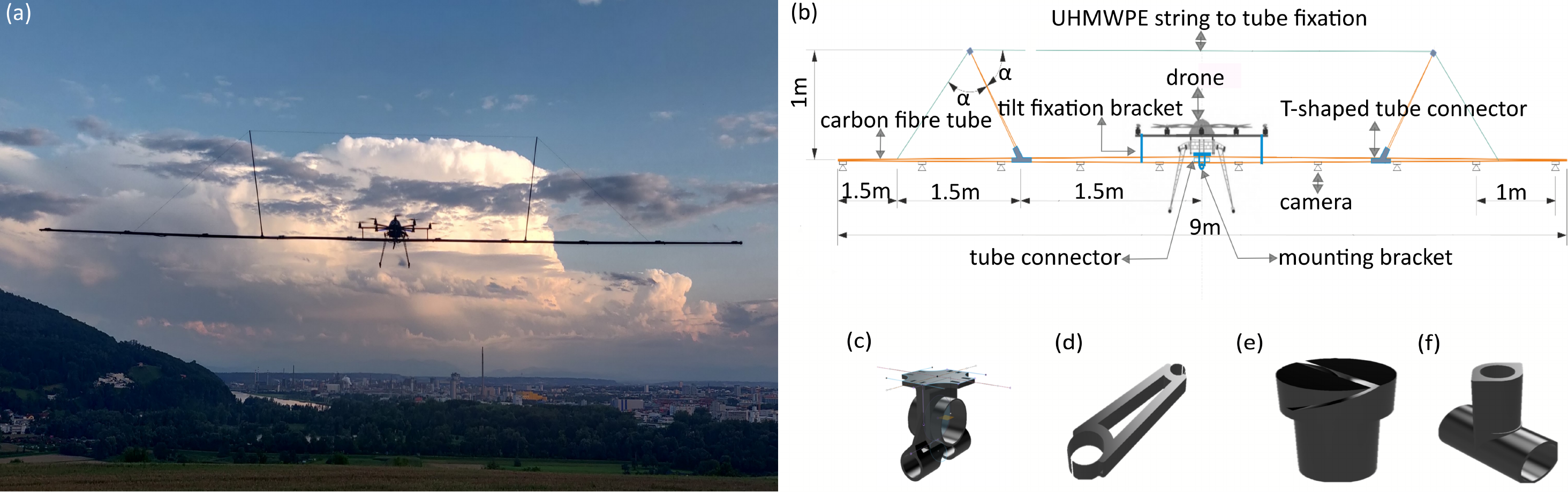}
	\caption{Drone prototype (a), design and construction of camera array (b), and custom-built 3D printed parts (c-f).}
	\label{FIG:2}
\end{figure*}
As illustrated in Fig. \ref{FIG:1}, our new payload captures multiple aerial images with a drone-operated 1D camera array. It samples the forest in parallel at flying altitude (\textit{h}) within the range of a synthetic aperture (\textit{SA}) that equals the size of the camera array. This results in a structured 4D light-field formed by image pixels which are represented as light rays in a 3D volume, as discussed in \cite{wetzstein2011computational,wu2017light}. With known camera intrinsics, camera poses, and a representation of the terrain (either a digital elevation model if available, as in \cite{schedl2021autonomous}, or a focal plane approximation if not, as in \cite{kurmi2020fast}), each ray's origin on the ground can be reconstructed. An occlusion-reduced integral image can be obtained by averaging all the rays that have the same origin. Depending on occlusion density, more or fewer rays of a surface point contain information of random occluders, while others contain the signal information of the target, as shown in Fig. \ref{FIG:1}. Therefore, integrating multiple rays (i.e., averaging their corresponding pixels) results in focus of the target and defocus of the occluders. This increases the probability of detecting the target reliably under strong occlusion conditions \cite{schedl2020search}. 

In practice, the acquired images are pre-processed for intrinsic camera calibration, image undistortion/rectification and are cropped to a common field of view. Furthermore, poses of all cameras at each instance in time have to be estimated. While GPS and IMU measurements enable real-time pose estimation (as in \cite{schedl2021autonomous}), computer vision techniques (e.g., multi-view stereo and structure-from-motion) applied to the recorded images (as in \cite{schedl2020search}) are more precise. The pre-processed images are finally projected to a common ground surface representation (digital elevation model or focal plane), using the cameras' intrinsic parameters and poses, and are averaged to the resulting integral image. Details on the concrete implementation of the prototype presented in this article are provided in Section \ref{SEC:P}.    
\section{Prototype}
\label{SEC:P}

Figure \ref{FIG:2}a illustrates our current prototype. The drone basis system is a MikroKopter OktoXL 6S12 octocopter. The custom built light-weight camera array is based on a truss design (Fig. \ref{FIG:2}b) which can handle the forces and vibrations  due to wind by safely distributing them over the entire structure.

The supporting tubes tilted outwards towards the outermost section at an angle \(\alpha\) are linked to the hollow carbon fibre tubes using a 3D printed T-shaped tube connector (Fig.\ref{FIG:2}f) and are subjected to axial loading through an ultra high molecular weight poly-ethylene (UHMWPE) string with a diameter of 0.9mm and 95daN ultimate load carrying capacity. The pre-tensioning of the UHMWPE string prevents the downward bending of the structure. A labyrinth like connection allows easy assembly of the 3D printed fixation tube (Fig.\ref{FIG:2}e) and the UHMWPE string.  The hollow detachable carbon fibre tubes with thin walled circular cross-section are commercially available Preston response match landing rods that are manufactured through filament winding. These tubes have high bending stiffness and therefore minimizes the structure deflection due to its own dead weight. The 3D printed drone mounting bracket shown in Fig.\ref{FIG:2}c is used to mount the structure on the drone and prevents the carbon fibre tube from spinning while the tilt fixation bracket in Fig.\ref{FIG:2}d increases the stability of the structure. All custom-built 3D printed connectors and brackets are made of flexible rubber-like thermoplastic polyurethane (TPU 95 Shore) to suppress crack formation and handle vibrations. 

\begin{figure*}[!t]
	\centering
		\includegraphics[width=\linewidth]{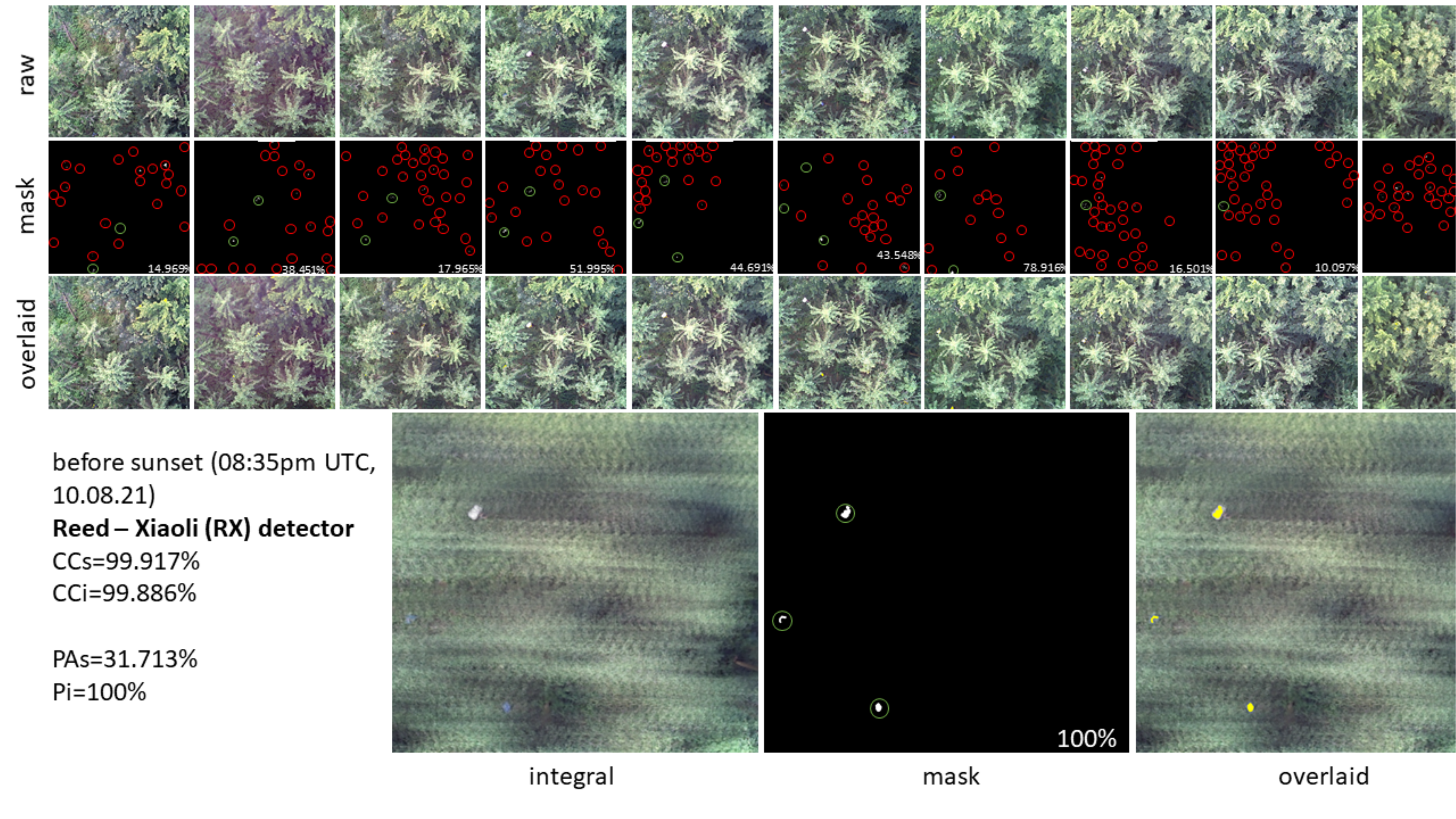}
	\caption{Color anomaly detection under bright light conditions (before sunset). Ten raw images recorded by camera array (top three rows): raw RGB images, RX anomaly mask, mask overlaid over raw image. Red and green circles indicate false and true positives. Numbers are precision values in \%. The corresponding integral image (raw, mask, and overlaid mask) is shown at the bottom row. Visible are three targets: two moving people and one static light source (largest cluster) for spatial reference.}
	\label{FIG:3}
\end{figure*}

The camera array carries ten light weight DVR pin-hole cameras (12g each), attached equidistant (1m) to each other on a 9m long detachable and hollow carbon fibre tube (700g) which is segmented into detachable sections of varying lengths and with a gradual reduction in diameter in each section from 2.5cm at the drone centre to 1.5cm at the outermost section. The cameras are aligned in such a way that their optical axes are parallel and pointing downwards. They record RGB images at a resolution of 1600x1200 pixels and RGB videos at a resolution of 1280x720px and 30fps to individual SD cards. All cameras receive power from two central 7.2V Ni-MH batteries and are synchronously triggered from the drone's flight controller trough a flat-band cable bus. To ensure stable flights and to avoid resonance oscillation with our payload, the drone's \textit{PID} controller had to be reconfigured. \textit{PID} stands for Proportional-Integral-Derivative which is a part of the flight control software that continuously reads the information provided by the drone's sensors and calculates how fast the motors must spin in order to retain the desired rotation speed for a stable flight. The \textit{P} controller changes the motor power proportional to the angle of inclination and the \textit{I} controller changes motor power continuously depending on the deflection angle and the time while the \textit{D} controller responds to any rapid changes in the sensor data. The \textit{PID} parameters were tuned to an \textit{I}-dominant state (Gyro \textit{P}: 100, Gyro \textit{I}: 255, Gyro \textit{D}: 10, all in a 0-255 range) for increased motor power to instantly compensate imbalances caused by external forces (such as wind) and to avoid oscillation of camera array's long lever. Supplementary videos 1 and 2 show test flights with default (causing resonance oscillation) and with tuned \textit{PID} parameters (ensuring stable flight).

For camera calibration, image undistortion, and rectification we apply OpenCV’s pinhole camera model (as explained in \cite{kurmi2018airborne,schedl2020search}). The undistorted and rectified images are cropped to a field of view of 41.10\degree and a resolution of 1024px × 1024px. Pose estimation is carried out using the general purpose structure-from-motion and multi-view stereo pipeline, COLMAP (\cite{schoenberger2016mvs,schonberger2016sfm}). Image back-projection and averaging is computed for a common focal plane at the ground surface level. Integral images are always rendered from the center perspective of the camera array. OpenGL deferred rendering is utilized here with  our GPU light field renderer implemented using Python, C, OpenGL, C++ and Cython integration\footnote{AOS source code, data, and publications are available at https://github.com/JKU-ICG/AOS/ .}. All computations are carried out offline (after landing), and require (without pose estimation but including RX detection) 896ms per integral image on an Intel(R) Core(TM) i5-6400 CPU @2.70 GHz (64GB RAM) with NVIDIA GeForce GTX 1070 GPU. Computer vision based pose estimation with COLMAP is slow and requires approximately 42s on our hardware. However, it becomes obsolete when being replaced by fast and precise online sensor measurements (such as Real Time Kinematic GPS). Furthermore, the CPU implementation of our RX detector is not performance optimized (requires 436ms / 896ms). A GPU implementation of it will lead to an additional speed-up.  
\section{AOS Enhanced Color Anomaly Detection}
Color anomaly detection finds pixels or clusters of pixels in an image with significant color differences in comparison to their neighbours.

 \begin{figure*}[!b]
	\centering
		\includegraphics[width=\linewidth]{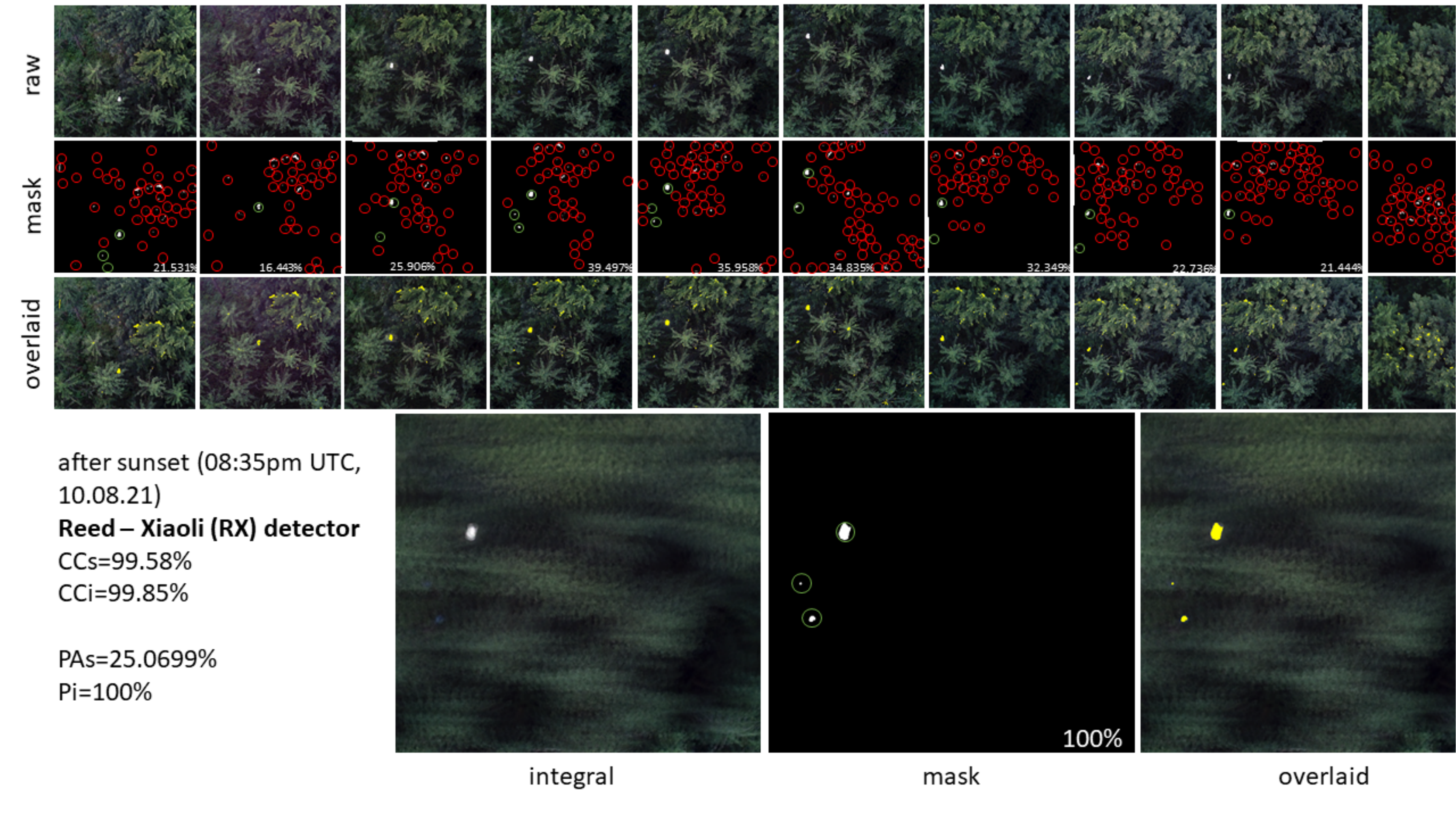}
	\caption{Color anomaly detection under dark light conditions (after sunset). Ten raw images recorded by camera array (top three rows): raw RGB images, RX anomaly mask, mask overlaid over raw image. Red and green circles indicate false and true positives. Numbers are precision values in \%. The corresponding integral image (raw, mask, and overlaid mask) is shown at the bottom row. Visible are three targets: two moving people and one static light source (largest cluster) for spatial reference.}
	\label{FIG:4}
\end{figure*}

Many color anomaly detection techniques are applied to hyper-spectral or multi-spectral images (\cite{manolakis2003hyperspectral}), with the Reed-Xiaoli (RX) anomaly detector being standard and most widely used (\cite{smetek2007finding}). The RX anomaly detector (\cite{reed1990adaptive}) characterizes the image background in terms of a covariance matrix and calculates the RX score based on the Mahalanobis distance between a test pixel \textit{r} and the background as follows:
 \begin{equation}
     \alpha_{RXD} (r) = (r-\delta)^T K^{-1}(r-\delta),
     \label{EQN:1}
 \end{equation}
where \textit{r} is the spectral 3-vector of the pixel under test, \(\delta\) is the spectral mean 3-vector of the background and \textit{K} is the 3x3-covariance matrix.
 
The RX scores are thresholded based on its cumulative probability distribution function. The threshold represents the RX score confidence beyond which the pixel under test is considered as an anomaly. The RX score confidence is set to maximize the detection of true positive pixels (i.e., maximum recall of the targets with a minimum one pixel) and minimize the detection of false positive pixels to attain the maximum pixel based precision value. 

\begin{table*}[H]     
\centering
\centering\caption{Comparison of color anomaly detection result between raw images and integral images before (SS-) and after (SS+) sunset.}\label{tbl1}
\begin{tabular}{  |P{0.8cm}|P{0.8cm}|P{0.8cm}|P{0.8cm}|P{0.8cm}|P{0.8cm}|P{0.8cm}|P{0.8cm}|P{0.8cm}|P{0.8cm}|P{0.8cm}|P{0.8cm}|P{0.8cm}|P{0.8cm}|}
 \hline
  flight & set & \multicolumn{10}{c|}{Ps}  & PAs & Pi \\
 \cline {3-12} 
  & no. & C0 & C1 & C2 & C3 & C4 & C5 & C6 & C7 & C8 & C9 &    &    \\
 \hline
 SS+ & 1 & 9.3 & 18.1 & 8.0 & 9.7 & 14.1 &  18.5 & 5.7 & 10.9 & 5.7 & 5.7  & 10.6 & 98.3   \\
 \hline
 SS- & 2 & 0 & 37.2 & 58.6 & 59.2 & 68.1 &  45.9 & 85.2 & 86.7 & 33.6 & 0  & 47.4 & 79.9   \\
 \hline
 SS- & 3 & 0 & 35.7 & 64.8 & 47.4 & 49.2 &  68.7 & 29.0 & 44.0 & 0 & 0  & 33.9 & 100   \\
 \hline
 SS+ & 4 & 84.5 & 81.8 & 67.2 & 83.5 & 94.9 &  89.3 & 94.1 & 84.9 & 72.2 & 0  & 75.2 & 98.8   \\
 \hline
 SS+ & 5 & 21.5 & 16.4 & 25.9 & 39.5 & 36.0 &  34.8 & 32.3 & 22.7 & 21.4 & 0  & 25.1 & 100  \\
 \hline
 SS+ & 6 & 34.9 & 98.0 & 96.5 & 95.9 & 64.3 &  84.7 & 64.7 & 96.6 & 100 & 92.2 & 82.8 & 100   \\
 \hline
 SS+ & 7 & 30.8 & 66.8 & 92.9 & 100 & 91.3 &  99.6 & 100 & 100 & 72.2 & 0  & 75.3 & 85.2   \\
 \hline
 SS+ & 8 & 75.4 & 100 & 78.0 & 94.1 & 99.0 &  100 & 79.1 & 100 & 95.1 & 35.4  & 85.6 & 100  \\
 \hline
 SS+ & 9 & 47.1 & 41.7 & 26.6 & 40.1 & 50.8 & 45.4 & 48.1 & 48.7 & 6.7 & 0  & 35.5 & 95.2   \\
 \hline
 SS- & 10 & 1.5 & 24.7 & 39.3 & 48.6 & 67.2 & 0.5 & 0 & 0 & 0 & 0  & 18.2 & 100   \\
 \hline
 SS- & 11 & 0 & 2.9 & 16.5 & 37.0 & 56.4 & 31.2 & 100 & 57.4 & 35.5 & 0  & 33.7 & 100   \\
 \hline
 SS- & 12 & 15.0 & 38.5 & 18.0 & 52.0 & 44.7 & 43.5 & 78.9 & 16.5 & 10.1 & 0  & 31.7 & 100   \\
 \hline
 SS- & 13 & 27.3 & 27.6 & 33.8 & 56.5 & 65.3 & 36.4 & 10.8 & 2.9 & 5.9 & 0  & 26.6 & 99.2   \\
 \hline
 SS- & 14 & 13.6 & 43.1 & 40.0 & 48.8 & 71.4 & 35.1 & 6.5 & 0 & 0 & 0  & 25.9 & 98.5   \\
 \hline
 SS- & 15 & 7.9 & 72.6 & 76.8 & 91.3 & 97.6 & 60.5 & 11.5 & 0 & 35.1 & 22.8  & 47.6 & 93.1   \\
 \hline
 SS- & 16 & 57.7 & 42.3 & 59.0 & 92.8 & 99.6 & 68.6 & 32.4 & 15.4 & 18.7 & 0  & 48.7 & 100   \\
 \hline
 SS- & 17 & 0 & 60.5 & 90.4 & 97.2 & 61.6 & 19.4 & 85.0 & 42.7 & 1.8 & 0  & 45.9 & 100   \\
 \hline
 SS+ & 18 & 20.4 & 25.6 & 24.9 & 25.4 & 15.0 & 16.2 & 23.7 & 14.2 & 2.0 & 14.4  & 18.2 & 91.4   \\
 \hline
 SS+ & 19 & 0 & 64.3 & 38.1 & 50.5 & 30.5 & 50.5 & 67.7 & 73.0 & 30.6 & 0  & 40.5 & 100  \\
 \hline
 SS+ & 20 & 9.5 & 38.8 & 17.8 & 6.0 & 24.4 & 68.0 & 81.4 & 26.9 & 0 & 0  & 27.3 & 100   \\
 \hline
  \multicolumn{12}{|c|}{} & \textbf{41.8} & \textbf{97.0}  \\
 \hline
\end{tabular}
	\label{TAB:1}
\end{table*}

Figures \ref{FIG:3} and \ref{FIG:4} present visual results of the RX anomaly detector applied to data sets captured with our drone prototype over dense mixed forest before and after sunset (performed in compliance with the legal flight regulations, before sunset+30min). We either apply the RX detector to the ten single images captured by the camera array individually, or to the corresponding integral image that results from registering and averaging the same ten single images. For both cases, optimal thresholds (i.e., one average threshold (\textit{CCs}) for all single images and one threshold (\textit{CCi}) for the integral image) are found as explained above (i.e., by maximizing true positive pixels and recall of the targets while minimizing false positive pixels). For all results, we use precision (ratio of true positive and true + false positive pixels in percent) as a quality metric. Here, \textit{PAs} is the precision average over all ten single images, and \textit{Pi} is the precision value of the corresponding integral image. Note, that the number of ground truth pixels (i.e., the number of target pixels under occlusion free conditions) are unknown. Note also, that the precision value is 0\% in case of no true positives (i.e., target is completely occluded or not within the field of view of the camera) but in the presence of false positives (i.e., wrong detections). For the case of no false positives but in the presence of true positives, the precision value is 100\%. 

As demonstrated in Figs. \ref{FIG:3} and \ref{FIG:4}, color anomaly detection benefits significantly from image integration. While strong occlusion causes many false positives but little true positives in raw images, false positives are almost entirely removed and true positives are significantly increased in the integral images. 

Table \ref{TAB:1} presents quantitative results of twenty data sets captured before (\textit{SS-}) and after (\textit{SS+}) sunset. Here, \textit{Ps} are the precision values of raw images (\textit{C0-C9}), \textit{PAs} the precision average over all ten images in each set, and \textit{Pi} the precision value of the corresponding integral image. On average, we achieve an improvement from 42\% to 97\% in precision when AOS is used in combination with color anomaly detection. We did not find a significant difference of this improvement when separating the low light SS+ sets (48\% vs. 97\%) and bright light SS- sets (37\% vs. 97\%). All data sets are available (\cite{rakesh}).  We discuss the underlying cause for this enhancement in Section \ref{SEC:D}.    
\section{Tracking Occluded Targets}
By making use of the camera array's video recording functionality and by extending the AOS image integration process to compute integral videos allows the application of a simple multi-object tracker to the RX anomaly mask time series at recording speed (30fps in our case). 

For tracking, we apply a blob detector to the binary RX anomaly masks which finds groups of connected pixels that differ in properties like color or brightness compared to the background. The association of the detections to the same targets depends on their motions, which is estimated using a Kalman filter. 

\begin{figure*}[!ht]
	\centering
		\includegraphics[width=\linewidth]{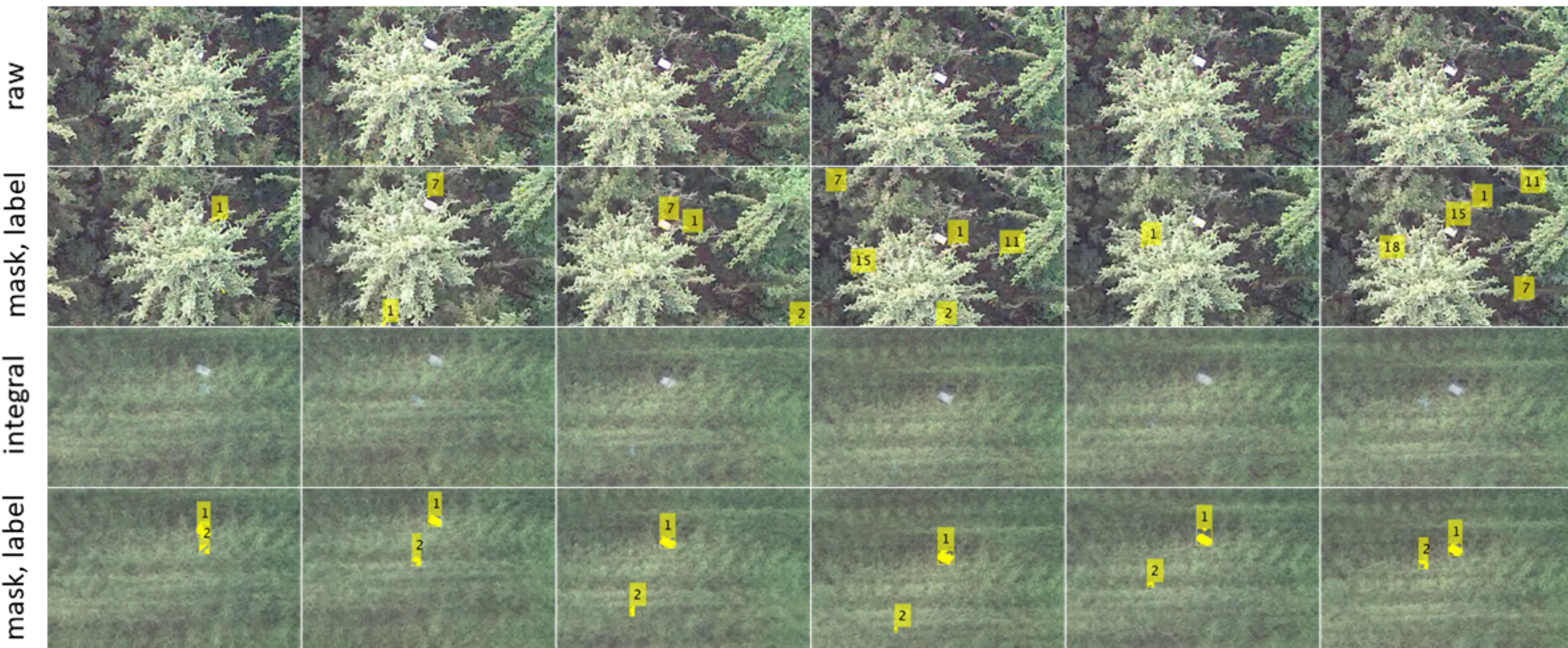}
	\caption{Motion based single-person tracking. Subset of single video frames (two top rows) and corresponding subset of integral frames (two bottom rows): raw RGB frames, RX anomaly mask and tracking labels overlaid. Visible are two targets: one moving person (label 2) and one static light source (label 1) for spatial reference.}
	\label{FIG:5}
\end{figure*}

\begin{figure*}[!ht] 
	\centering
		\includegraphics[width=\linewidth]{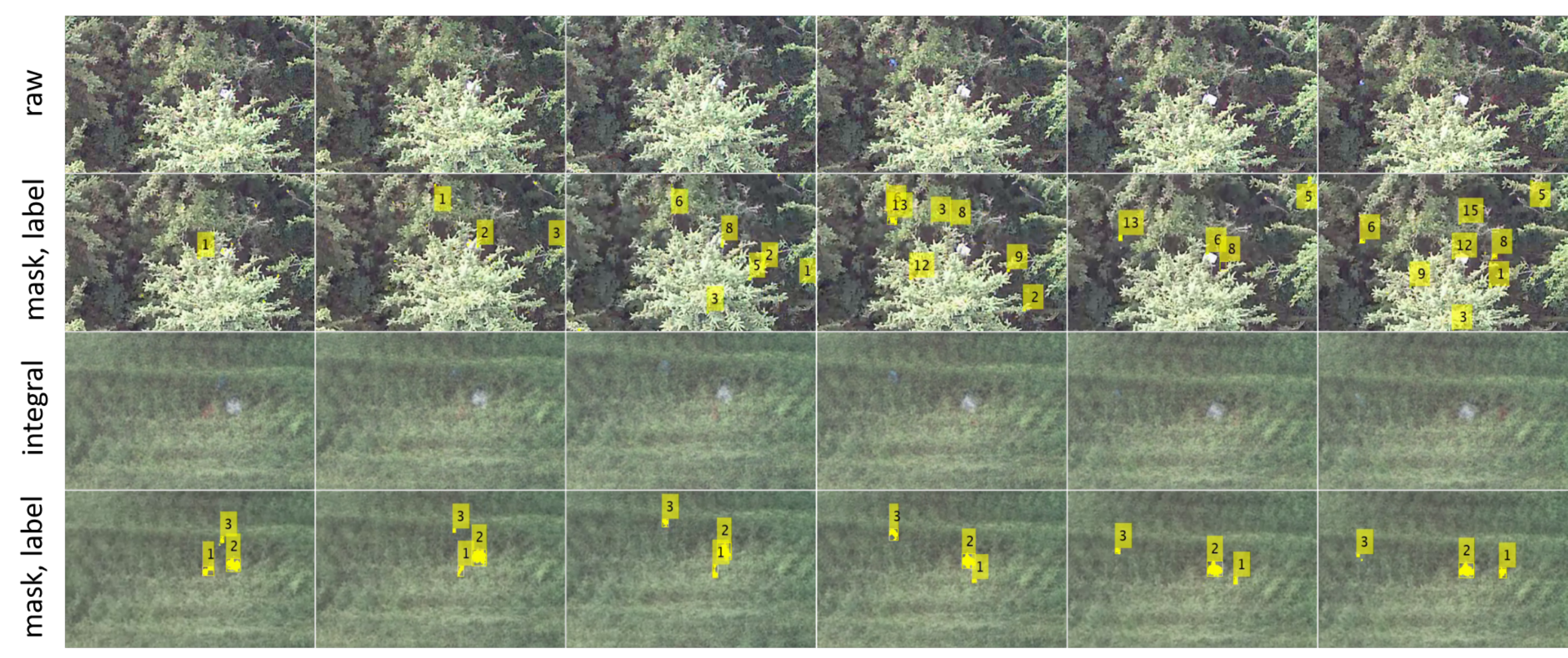}
	\caption{Motion based multi-person tracking. Subset of single video frames (two top rows) and corresponding subset of integral frames (two bottom rows): raw RGB frames, RX anomaly mask and tracking labels overlaid. Visible are three targets: two moving people (labels 1 and 3) and one static light source (label 2) for spatial reference.}
	\label{FIG:6}
\end{figure*}

As illustrated in Figs. \ref{FIG:5} and \ref{FIG:6}, object tracking in integral videos performs significantly better than in regular video recordings. While many false positive targets are detected, the true positive targets are not detected or are not consistently tracked throughout regular (single) video frames. However, few (none, in the examples shown in Figs. \ref{FIG:5} and \ref{FIG:6}) false positive targets are detected and all true positive targets are mostly consistently tracked throughout the integral video frames. Supplementary videos 3 and 4 show these examples in motion. 

The reason for the improved tracking quality is clearly the enhanced color anomaly detection in the integral video frames.

\begin{figure*}[!ht]
	\centering
		\includegraphics[width=\linewidth]{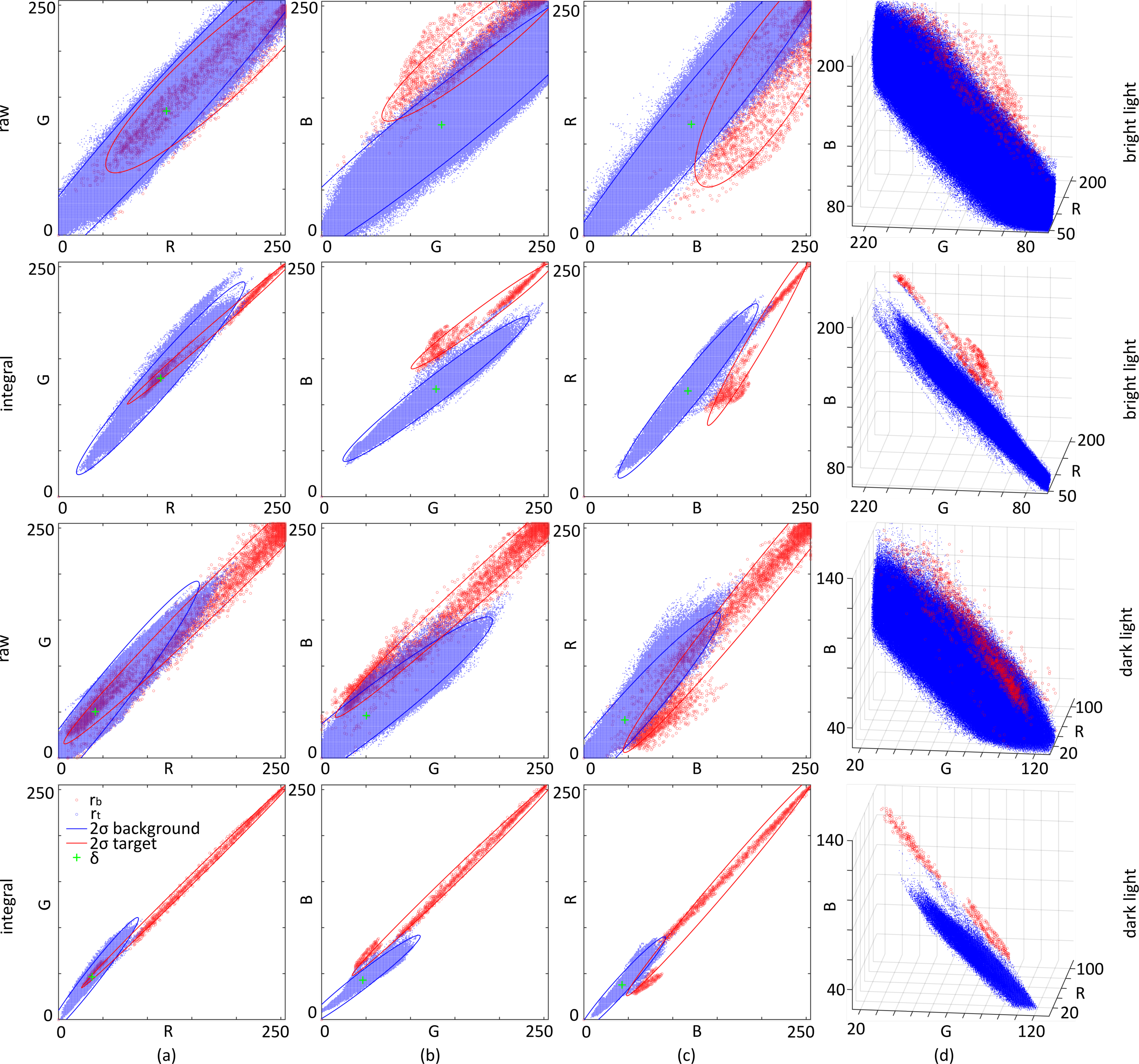}
	\caption{RGB values of target- ($r_t$, red circles) and background- ($r_b$, blue points) pixels for all raw images and resulting integral image of the two example sets shown in Figs. \ref{FIG:3} (top two rows) and \ref{FIG:4} (bottom two rows). The spectral mean background vectors ($\delta$, green crosses) and $2 \sigma $ ellipses (enclosing 95\% of the pixel data) generated from the eigen vectors and eigen values of the covariance matrices (blue and red ellipses for background and target data respectively) are illustrated. RGB space is projected to two principle axes in (a-c), and is 3D rendered, zoomed, and cropped to show details in (d).}
	\label{FIG:8plot}
\end{figure*}
\section{Discussion}
\label{SEC:D}
As shown in Eqn.\ref{EQN:1}, the RX anomaly detector (\cite{reed1990adaptive}) characterizes the image background in terms of a covariance matrix \textit{K} and calculates the RX score based on the Mahalanobis distance between a test pixel \textit{r} and the spectral mean vector of the background \(\delta\).

\begin{figure*}[!ht]
	\centering
		\includegraphics[width=\linewidth]{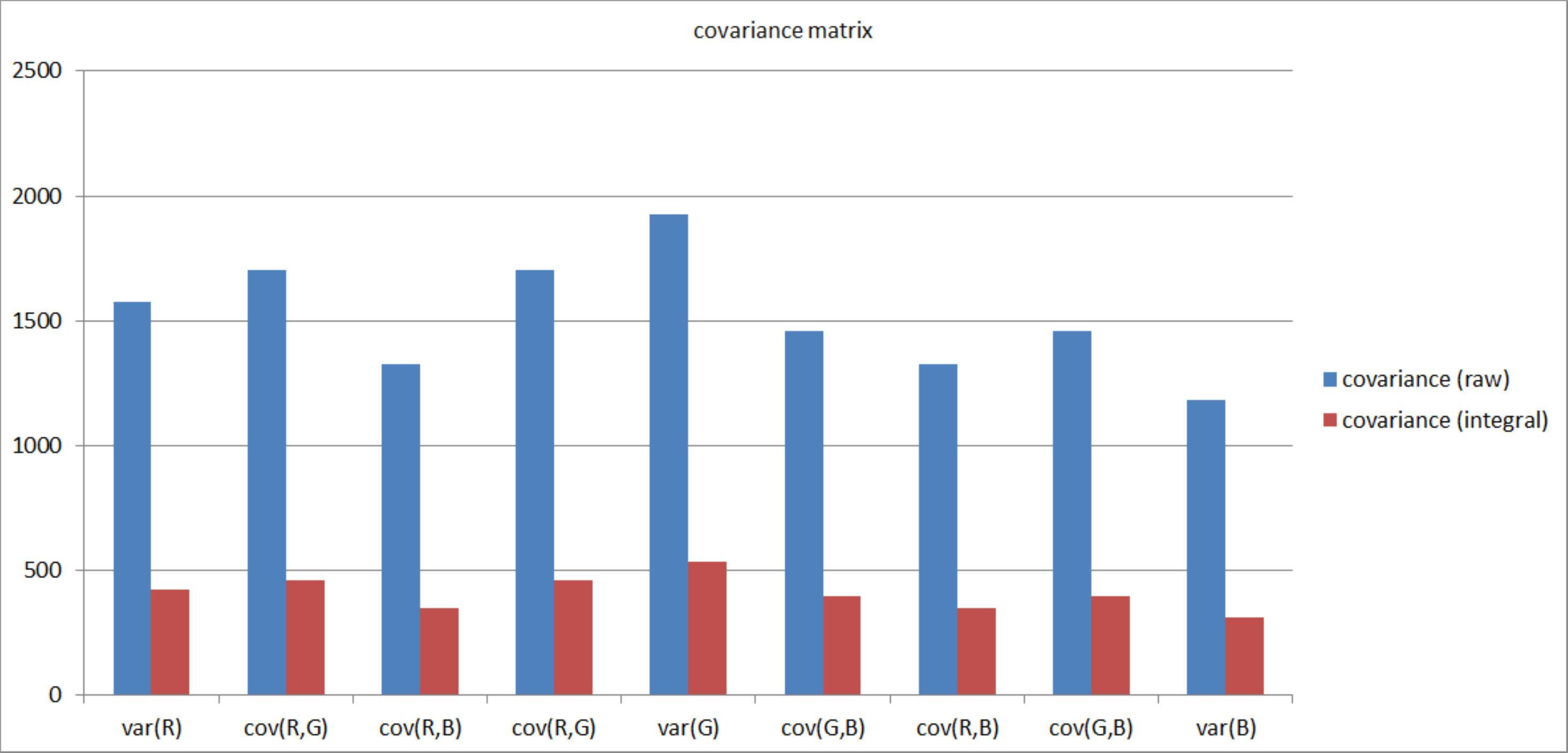}
	\caption{Comparison of 3x3 coefficients in covariance matrices \textit{K} for raw images and integral images (average over all images and all data sets presented in Tab. \ref{TAB:1}).}
	\label{FIG:9}
\end{figure*}

Figure \ref{FIG:8plot} visualizes scatter plots of RGB pixel values for all raw images and resulting integral image of the two example sets shown in Figs. \ref{FIG:3} and \ref{FIG:4}. Clearly, image integration decreases variance and co-variance, which leads to better clustering and consequently to enhanced separation of target and background pixels. This is evident by the integral images' shrunk $2 \sigma $ ellipses that results from the covariance matrices' decreased eigen vectors and coefficients. The coefficients of \textit{K} are decreased significantly (by a factor of 3 - 4 in our examples) for integral images when compared to the coefficient of \textit{K} in raw images (as shown in Figure \ref{FIG:9}, where we plot the average of coefficients for all 20 data sets as in Table \ref{TAB:1}). As a result, the distance between the two spectrally different clusters (background and targets) increases in integral images, which can then be well segmented using the Mahalanobis distance. Since the RX detector multiplies with the inverse of \textit{K} (Eqn. \ref{EQN:1}), lower coefficients lead to higher RX scores. 
 
AOS is an effective wide-synthetic-aperture aerial imaging technique that can also be considered as a variation of the standard signal averaging theorem (as explained in \cite{kurmi2019statistical}). In contrast to signal averaging, noise (occluders, such as vegetation in our case) is highly correlated (i.e., correlated background pixels of raw images shown in Fig. \ref{FIG:8plot}, raw). By shifting focus computationally towards the targets and by averaging multiple raw images accordingly leads to a strong and more uniform point-spread of out-of-focus occluders. This improves the signal-to-noise ratio between background and targets (as can be seen in Fig. \ref{FIG:8plot}, integral). 



\section{Conclusion and Future Work}
While detecting and tracking moving targets through foliage is difficult (and often even impossible) in regular aerial images or videos, it becomes practically feasible with image integration -- which is the core principle of Airborne Optical Sectioning. We have already shown that classification significantly benefits from image integration (\cite{schedl2020search}). In this work we demonstrate that the same holds true for color anomaly detection. This finding together with the implementation of an initial drone-operated camera array for parallel synthetic aperture aerial imaging allows presenting first results on tracking moving people through dense forest. Besides people, other targets (e.g., vehicles or animals) can be detected and tracked in the same way. This might impact many application domains, such as search and rescue, surveillance, border control, wildlife observation, and others.

The utilized RX color anomaly detector and the applied combination of blob detection and Kalman filter for tracking are only implementation examples that serve a proof-of-concept. They can be replaced by more advanced techniques. However, we believe that our main finding (i.e., anomaly detection and tracking benefit significantly from image integration) will still hold true. 

Color anomaly detection is clearly limited to detectable target color. In our experiments, targets were colored in white, black, blue, and red. Greenish color would have most likely not been detected. A combination of color (RGB), thermal (IR), and time (motion itself) channels for anomaly detection might result in further improvements. This has to be investigated in future. Furthermore, the implications of parallel-sequential sampling strategies and other sampling devices, such as re-configurable drone swarms instead of camera arrays with a fixed sampling pattern, have to be explored.  
\section*{Acknowledgements}
The camera array structure was designed and constructed by JKU's Institute of Structural Lightweight Design. This research was funded by the Austrian Science Fund (FWF) under grant number P 32185-NBL, and by the State of Upper Austria and the Austrian Federal Ministry of Education, Science and Research via the LIT – Linz Institute of Technology under grant number LIT-2019-8-SEE-114.

\printcredits

\bibliographystyle{main}

\bibliography{main}


\end{document}